\documentclass[11pt, a4paper, copyright]{perplexity}

\usepackage[authoryear, sort&compress, round]{natbib}
\usepackage{graphicx}
\usepackage{multirow}                                                                               
\usepackage{float}                                                                                   
\usepackage{adjustbox}
\usepackage{tablefootnote}

\uselogo{} 

\title{Diffusion-Pretrained Dense and Contextual Embeddings}

\author[*,1]{Sedigheh Eslami}
\author[*,1]{Maksim Gaiduk}
\author[*,1]{Markus Krimmel}
\author[*,1]{Louis Milliken}
\author[*,1]{Bo Wang}
\author[1]{Denis Bykov}

\affil[*]{Equal contributions}
\affil{Perplexity AI}

\begin{abstract}

In this report, we introduce \pplxfamily, a family of multilingual embedding models that employ multi-stage contrastive learning on a diffusion-pretrained language model backbone for web-scale retrieval.
By leveraging bidirectional attention through diffusion-based pretraining, our models capture comprehensive bidirectional context within passages, enabling the use of mean pooling and a late chunking strategy to better preserve global context across long documents.
We release two model types: \pplx for standard retrieval, and \pplxcontext for contextualized embeddings that incorporate global document context into passage representations.
\pplx achieves competitive performance on the MTEB(Multilingual, v2), MTEB(Code), MIRACL, BERGEN, and ToolRet retrieval benchmarks, while \pplxcontext sets new records on the ConTEB benchmark.
Beyond public benchmarks, \pplx demonstrates strong performance on our internal evaluation suite, focusing on real-world, large-scale search scenarios constructed from 1B production web pages. These results validate the models' effectiveness in production environments where retrieval quality and efficiency are critical at scale.
\end{abstract}

\begin{document}

\maketitle

\section{Introduction}

Dense textual embeddings represent texts as points in a continuous vector space in which distances capture meaningful semantic relationships. Embeddings are a crucial part of search systems, as they map queries and documents into a shared semantic space, in which information can be retrieved efficiently via approximate nearest neighbor search.
The recent development of large language models has increasingly shifted embedding model training toward employing pretrained decoder-only LLMs to leverage their pre-existing knowledge and improve embedding quality \citep{qwen3-embedding, jiang2024repurposing, lee2025gemini}.
We investigate diffusion-based language models~\citep{austin2021d3pm, nie2025llada} as an alternative paradigm. Diffusion language models employ transformer encoders with bidirectional attention, enabling more comprehensive context modeling compared to causally masked autoregressive models~\citep{zhang2025diffusion}.
This architectural difference is particularly advantageous for retrieval tasks, where capturing the global document context is essential.

This report introduces \pplxfamily\footnote{\url{https://huggingface.co/collections/perplexity-ai/pplx-embed}}, a family of multilingual text embedding models employing multi-stage contrastive learning on a diffusion-pretrained language model backbone for web-scale retrieval.
Continued pretraining via a diffusion objective converts a causally masked LLM backbone into a bidirectional encoder.
Further contrastive training on large-scale question-document pairs, as well as triplet data, aligns the embedding space geometry with semantic similarity. We release two model types: \pplx for standard retrieval and \pplxcontext for encoding passages with respect to document-level context.
Both models are released in 0.6B- and 4B-parameter scales.
Notably, our models are not instruction-tuned, eliminating the need for users to maintain instruction prefixes.

The \pplxfamily family utilizes native quantization-aware training and outputs INT8 embeddings by default, achieving competitive performance while offering significantly improved efficiency. On MTEB(Multilingual, v2)~\citep{enevoldsen2025mmteb,muenninghoff2023mteb}, \pplx-4B achieves an average \ndcgat{10} of 69.66\% with INT8 quantization, matching or exceeding leading models such as Qwen3-Embedding-4B (69.60\%) and gemini-embedding-001 (67.71\%).
The \pplx-4B model also demonstrates competitive performance on the MTEB(Code) and MIRACL benchmarks~\citep{miracl}. For contextual retrieval on ConTEB~\citep{conteb}, \pplxcontext-4B outperforms state-of-the-art contextual models such as voyage-context-3\footnote{\url{https://blog.voyageai.com/2025/07/23/voyage-context-3/}}~(79.45\%) and Anthropic Contextual\footnote{\url{https://www.anthropic.com/engineering/contextual-retrieval}}~(72.4\%) with an average \ndcgat{10} of 81.96\%. On ToolRet~\citep{shi2025retrieval}, \pplx-4B achieves 44.45\% average \ndcgat{10}, surpassing larger 7B models including NV-Embed-v1~\citep{nv-embed}~(42.71\%) and GritLM-7B~\citep{muennighoff2025generative}~(41.13\%). This enables efficient pre-filtering of relevant tools from large API corpora, reducing context usage.
We demonstrate that \pplx consistently outperforms BGE-M3~\citep{chen2024bgem3} and Qwen3-Embedding on the large-scale BERGEN~\citep{rau2024bergen} RAG benchmark, with \pplx-0.6B beating the larger Qwen3-Embedding-4B model on three of the five question-answering tasks. 
We further describe internal benchmarks for assessing embedding models as first-stage retrievers at web-scale and show superior performance of \pplx in comparison to Qwen3-Embedding and BGE-M3.

\section{PPLX Embedding}
\begin{figure}[ht]
    \centering
    \includegraphics[width=0.85\linewidth,trim={0.3cm 0.67cm 1.34cm 0.25cm},clip]{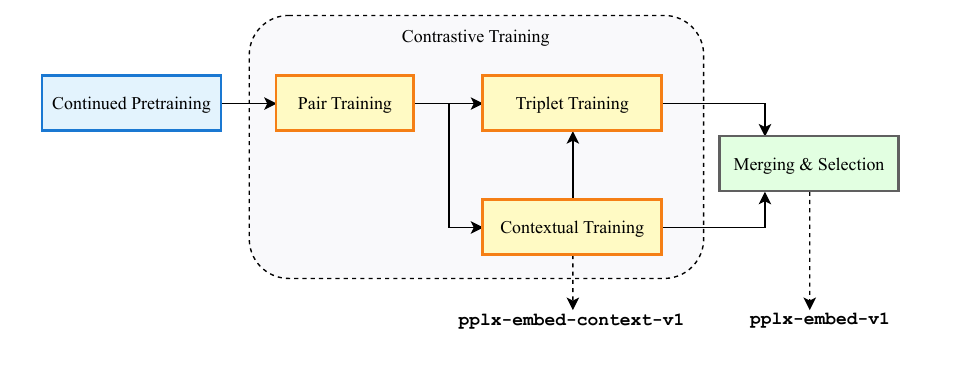}
    \caption{Training pipeline of \pplx and \pplxcontext.}
    \label{fig:pplx-pipeline}
\end{figure}

This section presents our training framework for learning high-quality embeddings for retrieval through a multi-stage curriculum. We combine four distinct training paradigms in a branched fashion, followed by a merging and selection stage. A schematic illustration of our training process is provided in Figure~\ref{fig:pplx-pipeline}. 

In the first training stage, described in Section~\ref{sec:continued-pretraining}, we perform continued pretraining of a decoder-only transformer on a diffusion objective, allowing it to use bidirectional self-attention. Subsequent training on query-document pairs (Section~\ref{sec:pair-training}) establishes basic semantic alignment of sequence-level embeddings. 
Following this, a contextual stage (Section~\ref{sec:contextual-training}) trains chunk-level embeddings, allowing the model to identify relevant passages within documents. The \pplxcontext model is obtained from this training stage. 
We perform triplet training with hard negatives (Section~\ref{sec:triplet-training}) on checkpoints obtained after pair and contextual training, refining boundaries between similar but non-relevant documents. For model merging, we employ Spherical Linear Interpolation~\citep{shoemake1985slerp} of the contextual model and the triplet checkpoints to obtain \pplx.

\subsection{Continued Diffusion Pretraining}
\label{sec:continued-pretraining}
Following the methodology of \citet{gong2025cpt}, we train two bidirectional diffusion language models via continued pretraining of existing autoregressive decoder-only backbones.
Considering the state-of-the-art performance of the Qwen3 family~\citep{yang2025qwen3}, we choose Qwen3-0.6B\footnote{\url{https://huggingface.co/Qwen/Qwen3-0.6B-Base}} and 4B\footnote{\url{https://huggingface.co/Qwen/Qwen3-4B-Base}} as our base models.

We disable causal attention masking and train the resulting transformer encoders to reverse a corrupting noise process. We adopt a continuous-time formulation~\citep{shi2024continuoustime} and an absorbing state process in which, at timestep $t \in [0, 1]$, each token has decayed to the absorbing \texttt{[MASK]} state independently with probability $t$. To represent the \texttt{[MASK]} state, we repurpose a rarely used token from the Qwen3 vocabulary. 
Following prior work~\citep{gong2025cpt,ye2025dream7b,nie2025llada}, we preserve the left-shift operation that is applied during autoregressive pretraining. Although we also performed experiments with annealing the causal attention mask~\citep{gong2025cpt}, we did not observe substantial performance improvements and did not pursue this technique further.

During training, we sample $t \sim \mathcal{U}(0.001, 1)$ for each input sequence independently and mask each token in the input sequence with probability $t$. 
We train our models via the standard evidence lower bound, which is given by the sum of token-wise cross entropies at masked positions, scaled by $1/t$. 

Half of our training data consists of English educational web pages from FineWeb-Edu~\citep{penedo2024finewebedu}, while the other half covers 29 other languages with data sourced from FineWeb2~\citep{penedo2025fw2} and FineWeb2-HQ~\citep{messmer2025multilingdatacomp}.
We train our models for 60,000 steps with a global batch size of 1024 and a sequence length of 4096. Thus, we perform pretraining on approximately 250 billion tokens of multilingual text data. Following \citet{nie2025scaling}, we truncate 1\% of the training sequences to a randomly chosen length to ensure that the models are exposed to varying sequence lengths. 
We use an AdamW optimizer~\citep{loshchilov2019adamw} with a warmup-stable-decay schedule and peak learning rates of $5\times10^{-4}$ and $3.16\times10^{-4}$ for the 0.6B and 4B models, respectively. For a more detailed description of hyperparameters, we refer the reader to Appendix~\ref{appendix:pretraining-details}. We leverage the resulting models exclusively for contrastive learning, evaluating them solely on embedding quality rather than generative performance.

\subsection{Pooling and Quantization}
To produce embeddings, we pool token-level representations extracted from the backbone model into a sequence-level representation. While recent embedding models based on decoder-only transformers~\citep{qwen3-embedding, tang2024pooling} typically employ last-token pooling, our bidirectional architecture allows the application of mean pooling. We propose a pooling method that natively combines mean pooling with quantization. Given token-wise embeddings $(\mathbf v_l)_{l=1}^L \in \R^{L \times d}$ for a sequence of length $L$, we define the sequence-level embedding as:
\begin{equation*}
    \floor*{127 \cdot \tanh\left(\frac{1}{L} \sum_{l=1}^L \mathbf v_l\right) + \frac{1}{2}}.
\end{equation*}
The resulting vector has integer entries in $\{-127, \dots, 127\}$, which are representable as signed 8-bit integers. 
We employ the quantization above not only during inference, but also during all contrastive training stages. We use straight-through gradient estimation~\citep{bengio2013straightthrough} to backpropagate through the non-differentiable rounding operation. The quantized embeddings are compared via their cosine similarity.

We also support binary quantization, which reduces the size of output embeddings by setting each entry of the mean-pooled embedding vector to $-1$ or $1$:
\begin{equation*}
\operatorname{bin}(x) = \begin{cases}
1 & \text{if } x \geq 0, \\
-1 & \text{otherwise.}
\end{cases}
\end{equation*}
While an embedding model could be trained using binary quantization with straight-through gradient estimation, we find that training-free post-hoc binarization can be applied with minimal performance loss.

\subsection{Pair Training}
\label{sec:pair-training}
Pair training represents the first contrastive learning stage, establishing foundational semantic alignment between queries and documents. The model learns to maximize the similarity of queries with their corresponding documents and minimize the similarity with unrelated ones. We employ an InfoNCE contrastive loss, which contrasts queries simultaneously against in-batch documents and other in-batch queries. 
Given a set of $N$ query-document pairs, we obtain corresponding embedding vectors~$\mathbf{q}_{i} \in \mathbb{R}^d$ and~$\mathbf{d}_{i} \in \mathbb{R}^d$ from our encoder for~$i=1,\dots, N$. For a temperature~$\tau > 0$, the loss is defined as:
\begin{equation*}
    \mathcal{L}^{\mathrm{pair}} =
    -\frac{1}{N} \sum_{i=1}^{N} \log \frac{e ^ {s(\mathbf{q}_i, \mathbf{d}_i)/ \tau}}{ e ^ {s(\mathbf{q}_i, \mathbf{d}_i)/ \tau} +  \sum_{j \neq i}^{N} m_{i}(\mathbf{d}_j) e ^ {s(\mathbf{q}_i, \mathbf{d}_j)/ \tau} + \sum_{j \neq i}^{N}m_{i}(\mathbf{q}_j) e ^ {s(\mathbf{q}_i, \mathbf{q}_j)/ \tau} },
\end{equation*}
with $m_{i}(\mathbf{x}) = \mathbbm{1}_{\{s(\mathbf{q}_i, \mathbf{x}) \le s(\mathbf{q}_i, \mathbf{d}_i) + 0.1\}}$ masking some terms, and~$s(\mathbf {q}_i, \mathbf {d}_i) = \frac{\mathbf{q}_i \cdot \mathbf{d}_i}{\Vert \mathbf{q}_i\Vert_2 \Vert \mathbf{d}_i \Vert_2}$ being the cosine similarity.
Inspired by \citet{qwen3-embedding}, we design the masking function $m$ to mitigate the effects of false negative samples. 
It compares the similarity of each in-batch negative to the query against that of the positive pair.
When a negative sample's similarity exceeds that of the positive pair by more than 0.1, indicating potential semantic relevance, and thus a likely false negative, the function masks its contribution, thereby preventing distortion of the learned representation space.

Pair training is conducted in three steps to gradually incorporate non-English data: 
first, the model is trained only on English, then on English and cross-lingual data, 
and finally on the entire pair dataset containing multilingual samples. More details on the training setup are provided in Appendix~\ref{appendix:contrastive-details}.

\subsection{Contextual Training}
\label{sec:contextual-training}
Contextual training is an approach for training embedding models on long documents divided into chunks
such that the embedding of each chunk retains contextual information from the whole document.
Given $N$ query-document pairs where each document contains $C$ chunks, $d_i = \{c_{ik}\}_{k=1}^C$, we compute embedding vectors $\mathbf{c}_{ik} \in \mathbb{R}^d$ for chunk $k$ from document $i$. We use a dual-objective loss function to capture local chunk-level semantics as well as global document-level representations. Inspired by \citet{conteb}, we define the local loss as a combination of the in-batch and in-sequence losses for chunks. For the in-sequence contrastive loss, the target (gold) chunk from a document is treated as the positive sample, and all remaining chunks from the same document are used as negatives. In contrast, for the in-batch loss, the gold chunk remains the positive sample, but the negatives are defined as all other chunks in the batch, including those from the same document. Using mean pooling and INT8 quantization to obtain the chunk-level embeddings, we define the sequence loss as:
\begin{equation*}
    \mathcal{L}^{\mathrm{seq}} = -\frac{1}{N} \sum_{i=1}^{N} \log \frac{e ^ {s(\mathbf{q}_i, \mathbf{c}_{i*})/ \tau}}{ \sum_{k=1}^{C} e ^ {s(\mathbf{q}_i, \mathbf{c}_{ik})/ \tau} },
\end{equation*}
with $\mathbf{c}_{i*}$ representing the embedding of the gold chunk. Furthermore, the in-batch loss is:
\begin{equation*}
\mathcal{L}^{\mathrm{batch}} = -\frac{1}{N} \sum_{i=1}^{N} \log \frac{e ^ {s(\mathbf{q}_i, \mathbf{c}_{i*})/ \tau}}{ \sum_{j=1}^{N} \sum_{k=1}^{C} e ^ {s(\mathbf{q}_i, \mathbf{c}_{jk})/ \tau}}.
\end{equation*}
The final local loss is then calculated by:
\begin{equation*}
\mathcal{L}^{\mathrm{local}} = \alpha \mathcal{L}^{\mathrm{seq}} + (1 - \alpha) \mathcal{L}^{\mathrm{batch}}.
\end{equation*}
In our experiments, we set $\alpha=0.2$.

For the global loss, we employ an InfoNCE objective to model query-document similarities. However, multiple queries within a batch may correspond to the same document, which would erroneously treat duplicate documents as negatives and introduce false negatives during training. To mitigate this, we identify and mask duplicate documents in the batch by comparing their hashes. Let $h(d)$ be a hash function mapping documents to identifiers (e.g., MD5). We introduce a duplicate indicator masking matrix~$M^\text{dup}$:
\begin{equation*}
M^{\text{dup}}_{ij} =
\begin{cases}
0, & \text{if } h(d_i) = h(d_j)  \text{ and } i \neq j, \\
1, & \text{otherwise.}
\end{cases}
\end{equation*}
Similar to the pair loss, we apply similarity threshold masking and include query-query negatives. Combining these with duplicate document masking, we define the global loss as:
\begin{equation*}
    \mathcal{L}^{\mathrm{global}} = 
    -\frac{1}{N} \sum_{i=1}^{N} \log \frac{e ^ {s(\mathbf{q}_i, \mathbf{d}_i)/ \tau}}{ \sum_{j=1}^{N} M_{ij}^\mathrm{dup} m_{i}(\mathbf{d}_j) e ^ {s(\mathbf{q}_i, \mathbf{d}_j)/ \tau} + \sum_{j \neq i}^{N}m_{i}(\mathbf{q}_j) e ^ {s(\mathbf{q}_i, \mathbf{q}_j)/ \tau}}.
\end{equation*}
For the total loss, \pplxcontext combines local and global losses with a scheduled weight $\beta$. We use a cosine schedule starting at $\beta=0.2$ with a final target value of $0.5$. The goal is for the model to first focus on learning local chunk-level semantics and gradually include document-level learning to mitigate forgetting of coarse document-level semantics. The total loss is defined as:
\begin{equation*}
\mathcal{L}^{\mathrm{context}} = \beta \mathcal{L}^{\mathrm{global}} + (1 - \beta) \mathcal{L}^{\mathrm{local}}.
\end{equation*}
We provide details of the training setup in Appendix~\ref{appendix:contrastive-details}.

\subsection{Triplet Training}
\label{sec:triplet-training}
Triplet training extends traditional pairwise contrastive learning by incorporating explicit hard negative examples alongside positive documents, enabling models to learn more discriminative embeddings through fine-grained relevance distinctions. Given a set of $N$ query-document triplets, we compute embeddings $\{\mathbf{d}^{h}_{ik} \in \R^d\}_{k=1}^K$ corresponding to the hard negatives of the query $\mathbf{q}_{i} \in \R^d$. The triplet contrastive InfoNCE loss is then formulated as:
\begin{equation*}
\mathcal{L}^{\mathrm{triplet}} = -\frac{1}{N} \sum_{i=1}^{N} \log \frac{e ^ {s(\mathbf{q}_i, \mathbf{d}_i)/ \tau}}{\sum_{j =1}^{N} e ^ {s\left(\mathbf{q}_i, \mathbf{d}_j\right)/ \tau} + \sum_{j=1}^N\sum_{k=1}^{K} e ^ {s\left(\mathbf{q}_i, \mathbf{d}^{h}_{jk}\right)/ \tau}}.
\end{equation*}
For more details on the training setup, we refer the reader to Appendix~\ref{appendix:contrastive-details}.

\subsection{Datasets for Contrastive Learning}
For contrastive training, we employ English, multilingual, and synthetic datasets. The final set contains 65.6\% English, 6.7\% cross-lingual, 1\% code, and 26.7\% multilingual samples from 60 different languages.
Contextual training is performed on the ConTEB training data, as well as data synthesized from the MLDR training set.
Triplet training uses considerably less but higher-quality data, spanning 12 datasets. Of this data, 92\% is English, 1\% is code, and 7\% consists of multilingual text covering 15 different languages.

All synthetic training data are generated using LLM-based synthesis with the Qwen3-30B-A3B-Instruct-2507 model. 
Inspired by \citet{qwen3-embedding}, our synthesis pipeline employs a two-stage persona-based approach to create diverse query-document pairs from web-scale corpora based on the top-5 relevant personas. 
For contextual training, we utilize a similar pipeline that generates synthetic queries for passages in a given document.

\section{Evaluations}

We evaluate our models across diverse benchmarks spanning multiple domains and languages.
Our evaluation suite focuses on retrieval tasks using public and in-house datasets.
In Section~\ref{sec:external-benchmarks}, we report results on the well-established MTEB~\citep{muenninghoff2023mteb,enevoldsen2025mmteb}, MIRACL~\citep{miracl}, ConTEB~\citep{conteb}, BERGEN~\citep{rau2024bergen}, and ToolRet~\citep{shi2025retrieval} benchmarks, providing a holistic measure of embedding quality across domains.
In Section~\ref{sec:internal-benchmarks}, we describe internal benchmarks designed as realistic indicators of the performance of embedding models in web-scale retrieval systems.

\subsection{Public Benchmarks}
\label{sec:external-benchmarks}

\paragraph{MTEB Multilingual.} The MTEB Multilingual v2 benchmark consists of 131 tasks, of which 18 measure retrieval performance across 146 languages. 
Table~\ref{tab:average-mteb-results} displays the storage efficiency of the embeddings (first column) alongside the average performance across these 18 tasks (second column).
Our 4B model, \pplx-4B, outperforms gemini-embedding-001 and closely rivals Qwen3-Embedding-4B in terms of average score while being substantially more storage-efficient.
Our 0.6B-parameter model, \pplx-0.6B, outperforms its Qwen counterpart. We provide per-task evaluation results and details on the evaluation approach in Appendix~\ref{appendix:mteb-eval-details}.

\begin{table}[tbp]
    \centering
    \small
    \caption{
    Storage efficiency (document embeddings per megabyte, higher is better) and average performance (\ndcgat{10}) on multilingual and code retrieval tasks.
    Upper group: models~$>$1B or unknown parameters; lower group: models~$<$1B parameters.
    The best score per group is in bold, second-best score underlined. 
    }
    \label{tab:average-mteb-results}
    \begin{tabular}{l|r|cc}
\toprule
Model & Docs/MB & MTEB(Multilingual, v2) & MTEB(Code) \\
\midrule
\pplx-4B (\textbf{INT8}) & \underline{390} & \textbf{69.66} & \underline{78.73} \\
\pplx-4B (\textbf{BIN}) & \textbf{3{,}125} & 68.22 & 78.11 \\
qwen3-embed-4B & 97 & \underline{69.60} & \textbf{80.07} \\
gemini-embedding-001 & 81 & 67.71 & 76.00 \\
text-embedding-3-large & 81 & 59.27 & 66.54 \\
\midrule
\pplx-0.6B (\textbf{INT8}) & \underline{976} & \textbf{65.41} & \textbf{75.85} \\
\pplx-0.6B (\textbf{BIN}) & \textbf{7{,}812} & 61.44 & 73.91 \\
qwen3-embed-0.6B & 244 & \underline{64.65} & \underline{75.42} \\
embed-gemma-0.3B & 325 & 62.58 & 68.76 \\
\bottomrule
\end{tabular}
\end{table}

\paragraph{MIRACL.}
To provide a more fine-grained illustration of our models' performance on specific languages, we present the evaluation results on the MIRACL benchmark, a subset of MTEB Multilingual, in Table~\ref{tab:miracl}. The MIRACL benchmark evaluates retrieval performance on 18 different languages across several scripts. On this benchmark, \pplx-0.6B performs particularly well, outperforming Qwen3-Embedding-0.6B on all language subsets. Even the binarized variant outperforms Qwen3-Embedding-0.6B in terms of average score. Notably, our 0.6B model exceeds the average score of our 4B model. On average, all \pplx models outperform text-embedding-3-large\footnote{\url{https://platform.openai.com/docs/guides/embeddings}} while trailing Qwen3-Embedding-4B and gemini-embedding-001. 

\begin{table}[tbp]
\centering
\caption{\ndcgat{10} on the MIRACLRetrievalHardNegatives task per language.}
\label{tab:miracl}
\addtolength{\tabcolsep}{-0.25em}
\resizebox{\textwidth}{!}{\begin{tabular}{l|r|c|cccccccccccccccccc}
\toprule
Model & Docs/MB & Avg & ar & bn & de & en & es & fa & fi & fr & hi & id & ja & ko & ru & sw & te & th & yo & zh \\
\midrule
\pplx-4B (\textbf{INT8}) & \underline{390} & 66.2 & 74.5 & 74.7 & 58.9 & 55.8 & 55.2 & 57.7 & 76.8 & 55.8 & 60.5 & 51.1 & 68.7 & 68.2 & 68.2 & \underline{71.4} & 78.8 & 74.6 & 80.2 & 61.3 \\
\pplx-4B (\textbf{BIN}) & \textbf{3{,}125}& 64.6 & 73.3 & 73.1 & 55.8 & 54.9 & 53.1 & 56.7 & 74.9 & 55.1 & 58.6 & 49.7 & 66.2 & 66.3 & 65.8 & 70.6 & 77.2 & 73.3 & 79.4 & 59.6 \\
gemini-embedding-001 & 81& \textbf{70.4} & \textbf{78.8} & \textbf{79.3} & \underline{60.7} & \underline{59.0} & \underline{57.5} & \underline{61.6} & \textbf{77.5} & \underline{56.3} & \textbf{65.1} & \underline{54.4} & \textbf{75.5} & \textbf{69.2} & \textbf{74.0} & \textbf{81.3} & \underline{80.7} & \textbf{81.3} & \textbf{89.1} & \textbf{66.4} \\
qwen3-embed-4B & 97 & \underline{69.5} & \underline{78.6} & \underline{78.3} & \textbf{63.0} & \textbf{59.6} & \textbf{58.9} & \textbf{62.2} & \textbf{77.5} & \textbf{60.2} & \underline{64.2} & \textbf{56.8} & \underline{72.5} & \underline{69.1} & \underline{71.7} & 68.9 & \textbf{84.1} & \underline{79.9} & \underline{80.3} & \underline{65.1} \\
text-embedding-3-large & 81 & 56.9 & 69.1 & 52.4 & 52.6 & 53.5 & 51.9 & 41.7 & 72.8 & 51.1 & 42.8 & 50.0 & 60.8 & 58.2 & 54.9 & 67.8 & 54.2 & 65.2 & 68.4 & 57.4 \\
\midrule
\pplx-0.6B (\textbf{INT8}) & \underline{976} & \textbf{68.6} & \textbf{77.9} & \underline{75.6} & \textbf{60.7} & \underline{57.5} & \textbf{60.1} & \textbf{60.1} & \textbf{76.6} & \textbf{59.9} & \textbf{61.2} & \textbf{55.0} & \textbf{70.8} & \textbf{70.3} & \textbf{69.5} & \textbf{74.3} & \textbf{78.6} & \textbf{78.7} & \textbf{82.1} & \underline{65.2} \\
\pplx-0.6B (\textbf{BIN}) & \textbf{7{,}812} & \underline{64.2} & \underline{73.9} & 71.8 & 54.7 & 54.1 & 53.7 & 55.9 & \underline{73.2} & 54.9 & 56.0 & 51.2 & 66.0 & \underline{67.5} & \underline{64.1} & \underline{70.0} & 75.4 & \underline{74.5} & \underline{80.3} & 58.7 \\
embed-gemma-0.3B & 325 & 62.3 & 72.8 & \textbf{76.0} & \underline{56.6} & \textbf{59.2} & \underline{55.9} & \underline{58.8} & 70.3 & \underline{58.0} & \underline{59.7} & 51.7 & \underline{68.8} & 64.1 & 61.8 & 62.6 & 63.1 & 62.8 & 48.5 & \textbf{68.0} \\
qwen3-embed-0.6B & 244 & 61.2 & 70.5 & 66.9 & 54.2 & 51.8 & 55.5 & 54.3 & 70.3 & 55.0 & 53.2 & \underline{52.4} & 63.1 & 62.0 & 61.0 & 49.0 & \underline{77.7} & 73.9 & 72.1 & 59.2 \\
\bottomrule
\end{tabular}}
\end{table}

\paragraph{MTEB Code.} The MTEB Code benchmark comprises 12 retrieval tasks across 15 different programming languages. We present the average performance in the third column of Table~\ref{tab:average-mteb-results}.
While our 4B model performs slightly worse than Qwen3-Embedding-4B, it outperforms text-embedding-3-large and gemini-embedding-001. Moreover, \pplx-0.6B outperforms its Qwen3 counterpart. Per-task scores are provided in Appendix~\ref{appendix:mteb-eval-details}.

\paragraph{ConTEB.}
We evaluate our models on the ConTEB~\citep{conteb} benchmark using the authors' evaluation framework
\texttt{cde\_benchmark}\footnote{\url{https://github.com/illuin-tech/conteb}}, a standardized toolkit for assessing chunk-level retrieval performance. When a contextual model is provided, the toolkit employs late chunking~\citep{gunther2024latechunking} for encoding chunks. In our evaluations, we perform quantization on the pooled chunk embeddings and compute cosine similarity between query embeddings and all document chunk embeddings, ranking chunks by relevance score. Our evaluations are conducted on eight diverse datasets from the ConTEB suite: SQuAD, MLDR, NarrativeQA, Football, COVID-QA, Geography, ESG Reports, and Insurance. Appendix~\ref{appendix:conteb-eval-details} provides more details of the evaluation process. In Table~\ref{tab:res-conteb}, we provide a comparison of the \ndcgat{10} metric for contextual and non-contextual models. Our results show that \pplxcontext-4B yields the best performance while \pplxcontext-0.6B ranks third, outperforming contextually trained ModernBERT-Large and Anthropic Contextual, but trailing the voyage-context-3 model.
\begin{table}[tbp]
\centering
\caption{Comparison of performance on ConTEB. Models above the line are non-contextualized; models below are contextualized. Dataset abbreviations: Covid (covid-qa), ESG (esg-reports), FB (football), Geo (geography), Ins (insurance), MLDR (mldr), NQA (narrative-qa), SQ (squad). *~indicates numbers taken from \citet{conteb}.}
\label{tab:res-conteb}
\adjustbox{max width=\textwidth}{%
\begin{tabular}{l|c|cccccccc}
\toprule
Model & Avg & Covid & ESG & FB & Geo & Ins & MLDR & NQA & SQ \\
\midrule
\pplx-4B (\textbf{INT8}) & 58.83 & 63.81 & 47.23 & 34.26 & 73.57 & 14.96 & 79.76 & 81.66 & 75.38 \\
\pplx-4B (\textbf{BIN}) & 57.91 & 62.29 & 46.83 & 31.92 & 72.16 & 15.75 & 79.10 & 80.82 & 74.38 \\
\pplx-0.6B (\textbf{INT8}) & 55.32 & 63.44 & 42.01 & 29.29 & 63.60 & 10.13 & 79.84 & 80.71 & 73.50\\
\pplx-0.6B (\textbf{BIN}) & 53.29 & 59.59 & 40.13 & 25.96 & 60.83 & 11.09 & 78.95 & 78.63 & 71.12 \\
qwen3-embed-0.6B & 49.36 & 49.10 & 34.72 & 23.15 & 58.11 & 12.65 & 76.27 & 74.02 & 66.85 \\
qwen3-embed-4B & 54.81 & 54.76 & 39.65 & 31.64 & 71.81 & 14.18 & 76.15 & 77.62 & 72.68 \\
\midrule
\pplxcontext-4B (\textbf{INT8}) & \textbf{81.96} & \textbf{62.16} & \textbf{62.40} & \underline{78.13} & \textbf{93.04} & \textbf{100} & \textbf{89.50} & \textbf{86.71} & \textbf{83.73} \\
\pplxcontext-4B (\textbf{BIN}) & \underline{80.46} & 59.60 & \underline{58.14} & 76.49 & 92.56 & \underline{99.69} & 88.90 & \underline{85.87} & 82.67 \\
\pplxcontext-0.6B (\textbf{INT8}) & 76.53 & 56.21 & 46.92 & 71.43 & 89.38 & 99.28 & 85.99 & 84.13 & 78.91 \\
\pplxcontext-0.6B (\textbf{BIN}) & 71.69 & 50.28 & 33.85 & 66.35 & 86.99 & 96.23 & 82.84 & 80.54 & 76.40\\
voyage-context-3 & 79.45 & 55.43 & 54.00 & \textbf{79.56} & \underline{92.85} & \textbf{100} & \underline{89.24} & 81.79 & \underline{82.70} \\
modernBERT-Large* & 75.6 & 56.0 & 43.1 & 63.9 & 90.7 & \textbf{100} & 88.7 & 81.3 & 80.9 \\
anthropic contextual* & 72.4 & \underline{60.7} & 34.8 & 53.9 & 89.4 & \textbf{100} & 85.4 & 77.7 & 77.1 \\
\bottomrule
\end{tabular}
}%
\end{table}

\paragraph{BERGEN.} To demonstrate the effectiveness of \pplx in large-scale RAG pipelines, we present evaluations on the BERGEN benchmark~\citep{rau2024bergen}. We index the KILT Wikipedia dump~\citep{petroni2021kilt}, consisting of $24.8$~million non-overlapping 100-word passages, using \pplx, Qwen3-Embedding, and BGE-M3. 
Following the recommendation of \citet{rau2024bergen}, we evaluate the five question-answering tasks that benefit most from retrieval augmentation: ASQA, HotpotQA, NQ, TriviaQA, and PopQA. For each question, the top-5 retrieved passages are presented to Qwen2.5-32B-Instruct\footnote{\url{https://huggingface.co/Qwen/Qwen2.5-32B-Instruct}}, which generates an answer. 
In Table~\ref{tab:bergen}, we report the match metric, measuring the proportion of generated answers that contain the corresponding ground-truth label. We observe that \pplx performs strongly across all tasks. The \pplx-4B model achieves the best results in three of the five tasks, outperforming Qwen3-Embedding-4B in four tasks. We also note that \pplx-0.6B outperforms Qwen3-Embedding-4B in three of the five tasks, highlighting its strong performance despite its small size. 
\begin{table}[tbp]
\caption{Match metric on the BERGEN benchmark with \texttt{Qwen/Qwen2.5-32B-Instruct} as a generator. No reranking is performed and answers are generated based on the top-5 retrieved passages. The standard retrieval prompt is prepended to queries for Qwen3-Embedding. See Appendix~\ref{appendix:bergen-eval-details} for details.}
\label{tab:bergen}
\centering
\small
\resizebox{\textwidth}{!}{\begin{tabular}{l|ccccc}
\toprule
Model & KILT-NQ & KILT-HotpotQA & KILT-TriviaQA & ASQA & PopQA \\
\midrule
\pplx-4B (\textbf{INT8}) & \textbf{67.7} & \textbf{51.9} & \textbf{91.9} & 71.5 & \underline{68.7} \\
\pplx-0.6B (\textbf{INT8}) & \underline{67.2} & \underline{51.6} & 91.0 & \underline{72.6} & \textbf{70.0} \\
qwen3-embed-4B & 67.1 & 50.2 & \underline{91.5} & \textbf{72.7} & 66.4 \\
qwen3-embed-0.6B & 63.0 & 47.2 & 88.3 & 66.9 & 63.9 \\
bge-m3 & 66.8 & 49.3 & 89.4 & 69.4 & 68.5 \\
\bottomrule
\end{tabular}}
\end{table}

\paragraph{Tool Search.}
We evaluate our models on the ToolRet benchmark~\citep{shi2025retrieval}, a comprehensive tool retrieval dataset consisting of 35 tasks across three categories: Web (19 tasks), Code (7 tasks), and Custom (9 tasks).
The benchmark contains diverse queries requiring models to retrieve relevant tools from a corpus of API documentation, with evaluation metrics including \ndcgat{10}, Precision@10, Recall@10, and Comprehensiveness@10.
As shown in Table~\ref{tab:toolret_results}, our \pplx-4B model achieves an average \ndcgat{10} of 44.45\%, ranking second overall among all evaluated models and demonstrating particularly strong performance on the Web category (42.07\% \ndcgat{10}).
Despite using INT8 quantization, our models remain competitive with larger full-precision baselines, with \pplx-0.6B achieving 43.05\% average \ndcgat{10} while being significantly more efficient.
Semantic retrieval significantly reduces context explosion by identifying relevant tools from large API corpora, enabling more efficient context management.

\begin{table*}[tbp]
\caption{Results on ToolRet benchmark. All metrics are @10. N=nDCG, P=Precision, R=Recall, C=Comprehensiveness. \pplx is evaluated with INT8 precision.}
\label{tab:toolret_results}
\centering
\scriptsize
\setlength{\tabcolsep}{2pt}
\resizebox{\textwidth}{!}{\begin{tabular}{l|cccc|cccc|cccc|cc}
\toprule
\multirow{2}{*}{\textbf{Model}} & \multicolumn{4}{c|}{\textbf{Web}} & \multicolumn{4}{c|}{\textbf{Code}} & \multicolumn{4}{c|}{\textbf{Custom}} & \multicolumn{2}{c}{\textbf{Avg}} \\
& N & P & R & C & N & P & R & C & N & P & R & C & N & C \\
\midrule
bm25 & 26.33 & 6.10 & 34.22 & 22.79 & 41.90 & 6.20 & 56.49 & 55.39 & 41.16 & 8.39 & 48.60 & 38.90 & 36.46 & 39.03 \\
gtr-t5-large & 24.37 & 5.27 & 31.64 & 21.26 & 36.76 & 5.33 & 47.42 & 45.92 & 42.04 & 8.48 & 50.84 & 40.00 & 34.39 & 35.73 \\
gte-base & 30.75 & 7.00 & 39.44 & 25.88 & 41.68 & 6.20 & 53.96 & 51.64 & 37.95 & 6.96 & 46.57 & 38.10 & 36.79 & 38.54 \\
bge-large & 30.03 & 7.01 & 39.28 & 25.63 & 41.53 & 6.00 & 52.76 & 51.18 & 43.90 & 8.31 & 51.79 & 42.24 & 38.49 & 39.68 \\
e5-mistral-7B & 31.07 & 7.65 & 41.30 & 27.04 & 44.97 & 6.66 & 58.95 & 56.79 & 40.88 & 7.91 & 49.35 & 38.35 & 38.97 & 40.73 \\
nv-embed-v1 & 31.51 & 7.74 & 40.52 & 26.74 & \textbf{47.92} & \underline{7.10} & \textbf{62.07} & \textbf{59.60} & 48.70 & \underline{10.07} & 57.69 & 43.88 & 42.71 & 43.41 \\
gte-qwen2-1.5B & \underline{37.53} & 9.31 & \underline{48.31} & \underline{30.95} & \underline{47.38} & \textbf{7.29} & \underline{61.12} & \underline{59.55} & \textbf{52.98} & \textbf{10.63} & \textbf{59.47} & \underline{45.68} & \textbf{45.96} & \textbf{45.39} \\
gritlm-7B & 36.58 & \underline{9.34} & 46.01 & 27.65 & 41.26 & 6.17 & 53.81 & 52.07 & 45.55 & 9.74 & 54.01 & 41.40 & 41.13 & 40.37 \\
\midrule
\pplx-0.6B & 35.84 & 8.19 & 45.26 & 30.31 & 45.52 & 6.62 & 59.07 & 57.32 & 47.79 & 9.24 & 55.33 & 45.24 & 43.05 & 44.29 \\
\pplx-4B & \textbf{42.07} & \textbf{9.89} & \textbf{52.55} & \textbf{36.42} & 41.61 & 6.20 & 54.96 & 53.12 & \underline{49.68} & 9.53 & \underline{57.77} & \textbf{45.98} & \underline{44.45} & \underline{45.17} \\
\bottomrule
\end{tabular}
}
\end{table*}

\subsection{Internal Benchmarks}
\label{sec:internal-benchmarks}

Public benchmarks are useful, but they do not fully capture web-scale retrieval challenges such as long-tail queries, noisy documents, and distribution shifts in production.
To benchmark performance in realistic deployment scenarios, we built PPLXQuery2Query~(PPLXQ2Q) and PPLXQuery2Doc~(PPLXQ2D), web-scale benchmarks with up to 115K real-world queries spanning a range of difficulty levels, evaluated against more than 30 million documents pooled from over 1 billion web pages.
This setup provides a more reliable estimate of recall and ranking performance under realistic corpus sizes and noise.

\paragraph{PPLXQuery2Query.}

We construct the PPLXQuery2Query benchmark from real search logs spanning five consecutive days from our production
search system.
Our key insight is that queries leading to the same destination URL exhibit semantic similarity, providing natural supervision for query-to-query retrieval without manual annotation.
The construction process proceeds as follows:
(1) apply Personally Identifiable Information (PII) detection to exclude any queries that could reveal user identity;
(2) collect query-URL pairs from search logs;
(3) group queries by destination URL, creating clusters of semantically related queries;
(4) filter clusters to retain only those with $\geq$2 queries and remove exact string duplicates; (5) within each cluster, designate the temporally first query as the evaluation query and the remaining queries as pseudo documents.
This yields a query set of 100K instances evaluated against document corpora of increasing size (240K, 1.2M, 2.4M), enabling analysis of scale-dependent retrieval performance.
Evaluation uses Recall@K (K $\in$ \{10, 20, 100\}), where a retrieved document is considered correct if it shares the same destination URL cluster as the query. 
We compute Recall@K for each query and report the mean across the query set in Table~\ref{tab:pplxq2q}.

\begin{table}[tbp]
\centering
\caption{Query-to-Query Retrieval Performance on PPLXQ2Q Benchmark. Models are grouped by size (separator line at 1B parameters).}
\label{tab:pplxq2q}
\adjustbox{max width=\textwidth}{%
\begin{tabular}{l|ccc|ccc|ccc}
\toprule
\multirow{2}{*}{Model\tablefootnote{Qwen models evaluated using their default prompt as specified in \url{https://huggingface.co/Qwen/Qwen3-Embedding-0.6B/blob/main/config_sentence_transformers.json}}} & \multicolumn{3}{c|}{Small (240K)} & \multicolumn{3}{c|}{Medium (1.2M)} & \multicolumn{3}{c}{Large (2.4M)} \\
& R@10 & R@20 & R@100 & R@10 & R@20 & R@100 & R@10 & R@20 & R@100 \\
\midrule
\pplx-4B (\textbf{INT8}) & \textbf{85.04} & \textbf{88.15} & \textbf{92.75} & \textbf{77.47} & \textbf{81.87} & \textbf{88.55} & \textbf{73.46} & \textbf{78.26} & \textbf{86.17} \\
\pplx-4B (\textbf{BIN}) & \underline{84.02} & \underline{87.28} & \underline{91.98} & \underline{76.44} & \underline{80.77} & \underline{87.69} & \underline{72.41} & \underline{77.23} & \underline{85.21} \\
qwen3-embed-4B & 80.90 & 84.57 & 90.21 & 72.36 & 76.96 & 84.90 & 67.90 & 73.02 & 81.96 \\
\midrule
\pplx-0.6B (\textbf{INT8}) & \textbf{82.68} & \textbf{85.97} & \textbf{91.01} & \textbf{75.05} & \textbf{79.31} & \textbf{86.40} & \textbf{71.05} & \textbf{75.75} & \textbf{83.86} \\
\pplx-0.6B (\textbf{BIN}) & \underline{80.25} & \underline{83.69} & \underline{89.01} & \underline{72.58} & \underline{76.83} & \underline{84.11} & \underline{68.60} & \underline{73.24} & \underline{81.42} \\
bge-m3 & 73.70 & 77.39 & 84.08 & 65.63 & 69.85 & 77.76 & 61.78 & 66.15 & 74.69 \\
qwen3-embed-0.6B & 68.71 & 73.15 & 81.54 & 59.31 & 64.12 & 73.64 & 55.07 & 59.86 & 69.82 \\
\bottomrule
\end{tabular}
}%
\end{table}

On the PPLXQ2Q benchmark, \pplx models achieve leading performance in all size categories.
On the Large corpus (2.4M target queries), \pplx-4B achieves 73.46\% R@10 and 86.17\% R@100 with INT8 quantization, surpassing Qwen3-Embedding-4B (67.90\% R@10, 81.96\% R@100) by 5.56 and 4.21 percentage points, respectively.
The binary quantized variant maintains strong performance with minimal degradation (72.41\% R@10, 85.21\% R@100), demonstrating only a 0.96--1.05 percentage point drop while still outperforming all competitors.
Similarly, \pplx-0.6B achieves 71.05\% R@10 and 83.86\% R@100, establishing substantial margins over BGE-M3 (61.78\% R@10, 74.69\% R@100) and Qwen3-Embedding-0.6B (55.07\% R@10, 69.82\% R@100). 

\paragraph{PPLXQuery2Doc.}

We construct the PPLXQuery2Doc benchmark as follows: 
(1) We create a high-quality query set using stratified sampling across four dimensions: query intent (Informational, Navigational, Transactional, Factual lookup, Exploratory), query form (Keyword-based, Natural language question, Telegraphic, Long-form verbose), query length (Short: 1–2 tokens, Medium: 3–11 tokens, Long: 12+ tokens), and language distribution to ensure multilingual coverage.
(2) For each query, we retrieve documents using four retrieval systems—BM25~\citep{bm25}, BGE-M3~\citep{chen2024bgem3}, Multilingual-e5-large-instruct~\citep{wang2024multilingual}, and Qwen3-Embedding-0.6B~\citep{qwen3-embedding}—over a corpus of 1 billion real-world web pages.
(3) The union of the results from all four systems forms a candidate pool of documents per query. 
(4) We assign Boolean relevance labels by thresholding Reciprocal Rank Fusion (RRF) scores that aggregate ranking signals from the four retrieval systems and our production search system (which is independent of \pplxfamily), ensuring robust relevance judgments through multi-system consensus.

The resulting benchmark comprises 15,000 queries, with 9,380 in English and 5,620 in other languages.
Unlike existing public benchmarks that focus on nDCG at small cutoff values, we focus on Recall@K to better reflect real-world cascade search systems where embedding models serve as the first-stage retrieval component.
We evaluate across three benchmark sizes: Small (15K queries, 7.5M corpus), Medium (15K queries, 15M corpus), and Large (15K queries, 30M corpus), reporting Recall@10, Recall@20, and Recall@100 across all benchmarks, with additional Recall@1000 reported for the Large benchmark. The corresponding results for English and Multilingual sets are reported in Tables~\ref{tab:q2d_english} and~\ref{tab:q2d}, respectively.

\begin{table}[tbp]
\centering
\caption{Query-to-Document Retrieval Performance on PPLXQuery2Doc Benchmark (English). Models are grouped by size (separator line at 1B parameters).}
\label{tab:q2d_english}
\adjustbox{max width=\textwidth}{%
\begin{tabular}{l|ccc|ccc|cccc}
\toprule
\multirow{2}{*}{Model} & \multicolumn{3}{c|}{Small (7.5M)} & \multicolumn{3}{c|}{Medium (15M)} & \multicolumn{4}{c}{Large (30M)} \\
& R@10 & R@20 & R@100 & R@10 & R@20 & R@100 & R@10 & R@20 & R@100 & R@1000 \\
\midrule
\pplx-4B (\textbf{INT8}) & \textbf{16.29} & \textbf{26.00} & \textbf{61.38} & \textbf{13.76} & \textbf{21.84} & \textbf{51.05} & \textbf{12.18} & \textbf{19.22} & \textbf{44.26} & \textbf{88.23} \\
\pplx-4B (\textbf{BIN}) & \underline{15.73} & \underline{25.05} & \underline{59.97} & \underline{13.30} & \underline{20.92} & \underline{49.52} & \underline{11.74} & \underline{18.40} & \underline{42.82} & \underline{87.13} \\
qwen3-embed-4B & 11.93 & 19.73 & 52.72 & 9.82 & 16.00 & 42.31 & 8.46 & 13.73 & 35.53 & 83.13 \\
\midrule
\pplx-0.6B (\textbf{INT8}) & \textbf{14.82} & \textbf{23.38} & \textbf{56.89} & \textbf{12.54} & \textbf{19.60} & \textbf{46.59} & \textbf{11.14} & \textbf{17.17} & \textbf{40.17} & \textbf{84.43} \\
\pplx-0.6B (\textbf{BIN}) & \underline{13.34} & \underline{21.31} & \underline{53.33} & \underline{11.13} & \underline{17.62} & \underline{42.94} & \underline{9.72} & \underline{15.27} & \underline{36.42} & \underline{80.71} \\
bge-m3 & 11.37 & 18.69 & 49.69 & 9.42 & 15.27 & 39.27 & 8.12 & 13.08 & 32.76 & 78.23 \\
qwen3-embed-0.6B & 10.50 & 17.42 & 48.02 & 8.59 & 14.04 & 37.55 & 7.42 & 11.97 & 31.10 & 77.90 \\
\bottomrule
\end{tabular}
}%
\end{table}
\begin{table}[tbp]
\centering
\caption{Query-to-Document Retrieval Performance on PPLXQuery2Doc Benchmark (Multilingual). Models are grouped by size (separator line at 1B parameters).}
\label{tab:q2d}
\adjustbox{max width=\textwidth}{%
\begin{tabular}{l|ccc|ccc|cccc}
\toprule
\multirow{2}{*}{Model} & \multicolumn{3}{c|}{Small (7.5M)} & \multicolumn{3}{c|}{Medium (15M)} & \multicolumn{4}{c}{Large (30M)} \\
& R@10 & R@20 & R@100 & R@10 & R@20 & R@100 & R@10 & R@20 & R@100 & R@1000 \\
\midrule
\pplx-4B (\textbf{INT8}) & \textbf{21.05} & \textbf{32.35} & \textbf{69.70} & \textbf{17.87} & \textbf{27.14} & \textbf{59.19} & \textbf{15.58} & \textbf{23.80} & \textbf{51.84} & \textbf{91.66} \\
\pplx-4B (\textbf{BIN}) & \underline{20.42} & \underline{31.65} & \underline{68.56} & \underline{17.17} & \underline{26.45} & \underline{57.73} & \underline{14.97} & \underline{23.05} & \underline{50.42} & \underline{90.67} \\
qwen3-embed-4B & 17.73 & 27.80 & 64.08 & 14.93 & 23.18 & 53.36 & 13.06 & 20.19 & 46.47 & 88.58 \\
\midrule
\pplx-0.6B (\textbf{INT8}) & \textbf{19.51} & \textbf{29.92} & \textbf{66.30} & \textbf{16.29} & \textbf{24.97} & \textbf{55.58} & \textbf{14.33} & \textbf{21.90} & \textbf{48.40} & \textbf{89.05} \\
\pplx-0.6B (\textbf{BIN}) & \underline{17.66} & \underline{27.44} & \underline{62.56} & \underline{14.78} & \underline{22.69} & \underline{51.59} & \underline{12.96} & \underline{19.73} & \underline{44.38} & \underline{85.61} \\
bge-m3 & 16.57 & 26.16 & 60.14 & 13.60 & 21.42 & 49.18 & 11.80 & 18.46 & 42.14 & 83.33 \\
qwen3-embed-0.6B & 16.23 & 25.31 & 59.40 & 13.45 & 20.78 & 48.45 & 11.68 & 17.97 & 41.55 & 84.27 \\
\bottomrule
\end{tabular}
}%
\end{table}

\pplx demonstrates strong performance across both the English and Multilingual variants.
On the Large corpus, \pplx-4B achieves 88.23\% Recall@1000 on English and 91.66\% on Multilingual, surpassing Qwen3-Embedding-4B (83.13\% and 88.58\%, respectively).
The binary quantized variants maintain competitive performance with minimal degradation (87.13\% and 90.67\%, respectively), demonstrating the effectiveness of our quantization-aware training approach.
Similarly, \pplx-0.6B achieves 84.43\% (English) and 89.05\% (Multilingual) with INT8 quantization, outperforming all sub-1B competitors by substantial margins.
These high recall rates at $K=1000$ validate the models' effectiveness as first-stage retrievers in multi-stage ranking pipelines, where maximizing recall is critical for downstream reranking performance.

\subsection{Effect of Binary Quantization}
Across all benchmarks, the performance loss from binary quantization is considerably higher for \pplx-0.6B, whose performance drops by roughly 2--4.4 percentage points, compared to \pplx-4B, which only loses up to 1.6 percentage points. The same trend is observed when comparing \pplxcontext-4B and \pplxcontext-0.6B, with the 0.6B model losing up to 5 percentage points and the 4B model losing about 1 percentage point when using binary quantization. Besides the difference in the number of parameters, our 4B models may be more resilient to binarization due to their output dimension of 2560, compared to 1024 for \pplx-0.6B. This allows our 4B models to preserve more information in their compressed representations, making them less susceptible to the information loss inherent in quantization.

\section{Diffusion vs. Autoregressive Pretraining}
\begin{figure}[t]
    \centering
    \includegraphics[width=0.8\linewidth]{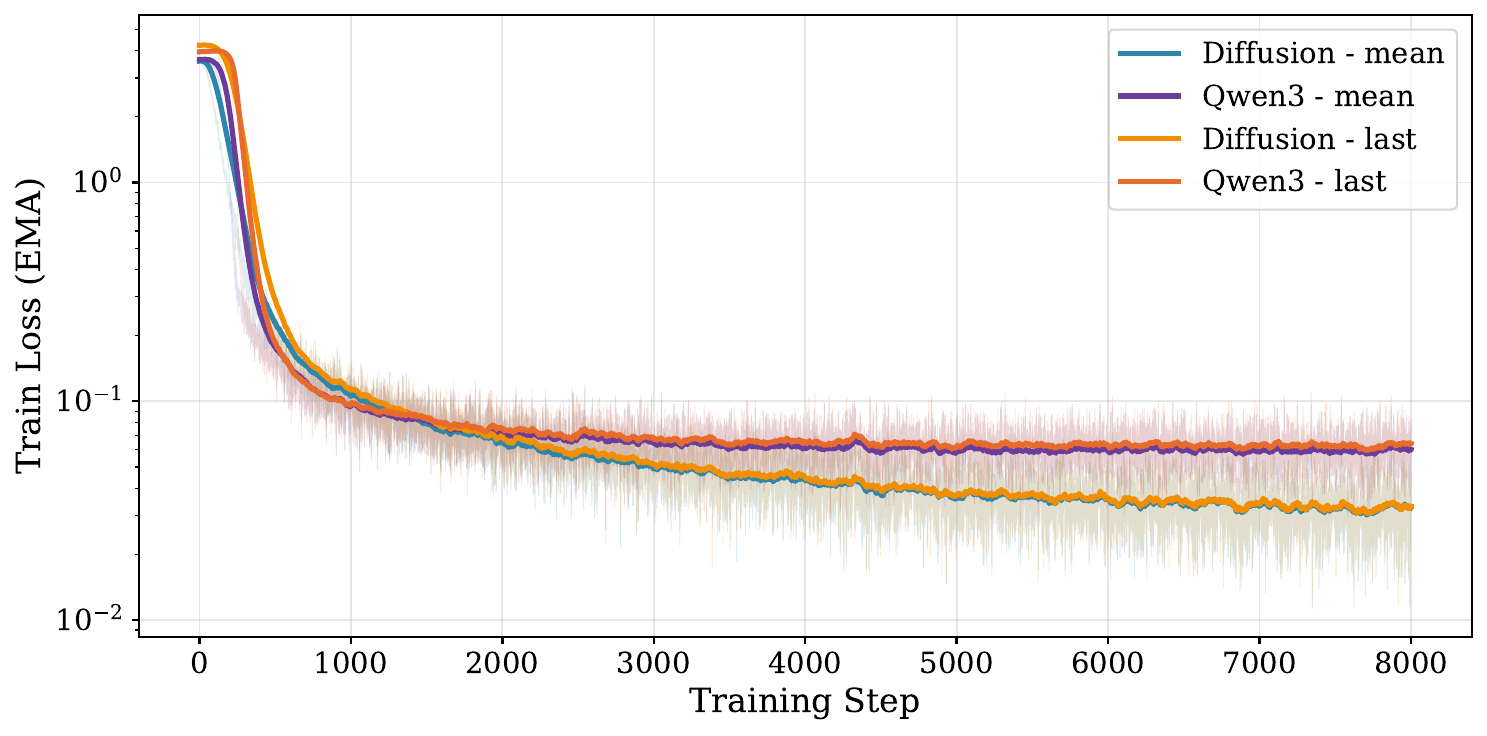}
    \caption{Smoothed training loss using exponential moving average with $\alpha=0.02$.}
    \label{fig:ablation_loss}
\end{figure}

In this section, we present an ablation study to demonstrate the effectiveness of diffusion pretraining and bidirectional attention. Starting from pretrained base models, we perform a small number of contrastive pair training steps and evaluate the performance of the resulting embedding models. We evaluate four configurations derived from two base models and two pooling strategies. We compare the causally masked Qwen3 base model (denoted as Qwen3) against a bidirectional backbone pretrained with a diffusion objective (denoted as Diffusion). For each backbone, we apply either mean pooling or last-token pooling. The Qwen3 base model remains causally masked during contrastive training, while the diffusion base model uses bidirectional attention throughout training.
We perform pair training on English data for less than one epoch. The training loss, presented in Figure~\ref{fig:ablation_loss}, serves as an initial indicator of model performance. We find that the model configurations using the bidirectional diffusion backbone achieve substantially lower loss values compared to those initialized with the causally masked Qwen3 model.

\begin{table}[t]
\centering
\footnotesize
\setlength{\tabcolsep}{4pt}
\caption{Effect of continued pretraining on selected tasks. Qwen3 refers to the causally masked Qwen/Qwen3-0.6B-Base model, mean and last indicate mean pooling and last-token pooling, respectively. Dataset abbreviations: Game (CQADupstackGamingRetrieval), Unix (CQADupstackUnixRetrieval), DBP (DBpedia), FEV (FEVER), FiQA (FiQA2018), HotQ (HotpotQA), MIR (MIRACLRetrieval), MSM (MSMARCO), NFC (NFCorpus), NQ (NaturalQuestions), SciD (SCIDOCS), SciF (SciFact).}
\label{tab:ablation}
\resizebox{\textwidth}{!}{\begin{tabular}{ll|cccccccccccc|c}
\toprule
Base & Pooling & Game & Unix & DBP & FEV & FiQA & HotQ & MIR & MSM & NFC & NQ & SciD & SciF & Avg \\
\midrule
\multirow{2}{*}{Qwen3}
  & last & 47.7 & \underline{33.2} & \textbf{31.2} & \textbf{72.1} & 26.6 & 51.7 & 39.3 & 28.3 & \textbf{30.9} & \underline{38.6} & \textbf{17.1} & 61.5 & \underline{39.9} \\
  & mean & 47.2 & 31.4 & 30.3 & \underline{68.1} & 27.0 & 50.5 & 37.8 & 27.3 & \underline{30.2} & 35.5 & \underline{16.8} & \underline{63.0} & 38.8 \\
  \midrule
\multirow{2}{*}{Diffusion}
  & last & \underline{48.6} & 32.0 & \underline{30.8} & 67.1 & \underline{28.8} & \underline{51.8} & \underline{41.4} & \underline{31.1} & 29.1 & 38.2 & 15.7 & 61.4 & 39.7 \\
  & mean & \textbf{49.9} & \textbf{33.4} & 30.2 & 66.5 & \textbf{31.1} & \textbf{54.2} & \textbf{41.5} & \textbf{31.3} & 29.6 & \textbf{39.0} & 16.5 & \textbf{64.6} & \textbf{40.6} \\
\bottomrule
\end{tabular}}
\end{table}

In Table~\ref{tab:ablation}, we further compare the performance of the four variants on English retrieval tasks consisting of the MTEB(En, v2) benchmark and the English subset of MIRACLRetrievalHardNegatives. We observe that the combination of mean pooling and diffusion pretraining provides improvements on a range of retrieval tasks, resulting in an increase of $\sim$1\ percentage point on average.
In addition to modestly improving benchmark performance, mean pooling is crucial for our contextual embedding training, as it enables computing many chunk-level representations from a single document.
\section{Related Work}

\paragraph{Diffusion Language Models.} Recent work has explored diffusion language models (DLMs) as an alternative to autoregressive models for text generation \citep{austin2021d3pm, nie2025llada, gong2025cpt}.
Moreover, \citet{zhang2025diffusion} conducted a systematic study of diffusion language models for text embeddings, demonstrating that bidirectional attention is crucial for encoding global context in long and complex documents. In this work, we study continued pretraining with diffusion language models and their performance in text retrieval.
As noted by \citet{austin2021d3pm}, training DLMs with an absorbing state process closely resembles masked language modeling~\citep{devlin2019bert} with randomly sampled masking ratios. In this sense, our approach extends a long line of work in which BERT models have been fine-tuned for retrieval~\citep{reimers2019sbert,gunther2023jinav2,chen2024bgem3}.

\paragraph{Contrastive Training of Text Embeddings.} InfoNCE-based contrastive learning is the predominant approach for training text embedding models, aligning representations of semantically similar texts while distinguishing dissimilar ones~\citep{reimers2019sbert, gao2021simcse, contriever, neelakantan2022textembedding, santhanam2022colbertv2}. Contemporary research focuses on improving sample quality through synthesis of pair data~\citep{qwen3-embedding, synthetic_data_1, synthetic_data_2}, data cleaning~\citep{thakur2025hard}, and hard negative mining techniques~\citep{chen2024bgem3}. Moreover, multi-stage contrastive training has emerged as an effective strategy for training embedding models~\citep{multi-stage-alibaba, nv-embed}. 

Recent work has integrated quantization into contrastive training, from Contrastive Quant in computer vision~\citep{fu2022contrastivequant} to EmbeddingGemma’s quantization-aware finetuning for weight-quantized variants~\citep{vera2025embeddinggemma}. \citet{huerga2025optimization-quantization} perform systematic evaluations of post-training quantization for RAG-based use cases. Our work, in contrast, targets quantization-aware training of embeddings for web-scale retrieval. 

\paragraph{Contextual Embeddings.} Several approaches have emerged for learning contextual embeddings. \citet{morriscontextual} train document embeddings to be contextualized with respect to their neighboring documents in the batch. In contrast, \citet{gunther2024latechunking} propose a training-free approach that captures context by processing the chunks in a single forward pass followed by chunk-level pooling. Furthermore, the Context-aware Text Embedding Benchmark (ConTEB)~\citep{conteb} was specifically introduced to evaluate models' ability to leverage global context. Additionally, the authors propose the in-sequence training approach for chunk-level contrast. Our work builds upon these foundations by employing multi-stage contrastive training with a specialized dual-loss objective, while maintaining compatibility with both MTEB and ConTEB evaluation protocols.
\vspace{-2mm}
\section{Conclusion}
This report presented the \pplxfamily model family, which builds on diffusion-based language models with bidirectional attention to train embedding models that better capture global document context. Our multi-stage training pipeline progressively shapes text representations for semantic alignment, contextualized chunk encoding relative to full documents, and fine-grained relevance distinctions. We provide four model variants—\pplx and \pplxcontext at 0.6B and 4B parameter scales—and show through extensive evaluation on public and internal web-scale benchmarks that they achieve strong retrieval performance while offering practical deployment benefits via native quantization-aware training that directly outputs INT8 and binary embeddings.


\vspace{-2mm}\section*{Acknowledgments}
We thank Ismail Gadzhiev and Alexander Pecheny for their contributions to the PPLX benchmarks and Lequn Chen, Svyatoslav Feldsherov, Kevin Hu, Nandor Licker, Sebastian Sepulveda, and Vladimir Zaytsev for their work on supporting \pplxfamily in inference. 
We also thank Tom Aarsen, Alvaro Bartolome, and Joshua Lochner for integrating \pplxfamily with sentence-transformers, text-embeddings-inference, and ONNX.

\clearpage

\bibliography{main}

\appendix
\clearpage
\begingroup
\renewcommand\thepart{}
\renewcommand\partname{}
\part{Appendix}
\endgroup

\section{Details on Continued Pretraining}
\label{appendix:pretraining-details}
\paragraph{Loss.} We perform diffusion pretraining via the standard evidence lower bound (ELBO) for a noise process in which tokens decay to an absorbing \texttt{[MASK]} state under a linear noise schedule. Given a sequence $x_0$ of length $L$, this forward process is modeled at timestep $t \in [0, 1]$ by $q(x_t | x_0)$, where each token of the noisy sequence $x_t$ has decayed to the absorbing state independently with probability $t$. We model the reverse process with our bidirectional transformer $p_\theta$. Indexing the $l$\textsuperscript{th} position of the sequence $x_t$ via the notation $x_t^l$, we formulate the loss as:
\begin{equation*}
    \mathcal{L}^{\mathrm{ELBO}}(x_0) = \mathbb{E}_{t \sim \mathcal{U}(0.001, 1)}\left[\frac{1}{t}\mathbb{E}_{q(x_t | x_0)}\left[-\sum_{l=2}^L \delta_{x_t^l, \texttt{[MASK]}} \log p_\theta(x_0^l \vert x_t)\right]\right].
\end{equation*}
Here, $\delta_{x_t^l, \texttt{[MASK]}}$ denotes the Kronecker delta, taking the value $1$ at masked positions and the value $0$ otherwise. We note that the transformer $p_\theta$ maintains a left-shift operation, and we do not prepend a \texttt{[BOS]} token to input sequences. Hence, we cannot make predictions for the first token of the masked sequence and the sum consequently runs only over the last $L-1$ positions.

\paragraph{Hyperparameters.} We perform pretraining using automatic mixed bfloat16 precision and the FlashAttention-2~\citep{dao2023flashattention2} implementation. The AdamW optimizer uses a weight decay of~$0.01$, and we set the exponential decay rates to $\beta_1 = 0.9$ and $\beta_2 = 0.98$. We apply gradient clipping based on the $\ell^2$ norm of the gradients. Since the appropriate clipping threshold depends on implementation details (e.g., loss scaling), we do not state it here. The learning rate is warmed up linearly over the first 6,000 steps and decayed in a cosine schedule over the last 12,000 steps. 

\paragraph{Data.} In Table~\ref{tab:pretraining-languages}, we list the composition of the pretraining data. We fix the prevalence of non-English languages according to their relative word count in the FineWeb2 corpus~\citep{penedo2025fw2}.
\begin{table}[htp]
\centering
\caption{Composition of pretraining data.}
\label{tab:pretraining-languages}
\begin{tabular}{cccc}
\toprule
Language & Script & Source & Prevalence \\
\midrule
eng & Latn & FineWeb-Edu & 50.00\% \\
rus & Cyrl & FineWeb2-HQ & 9.22\% \\
cmn & Hani & FineWeb2-HQ & 8.51\% \\
jpn & Jpan & FineWeb2-HQ & 5.19\% \\
deu & Latn & FineWeb2-HQ & 4.11\% \\
spa & Latn & FineWeb2-HQ & 4.10\% \\
fra & Latn & FineWeb2-HQ & 3.46\% \\
ita & Latn & FineWeb2-HQ & 2.18\% \\
por & Latn & FineWeb2-HQ & 1.72\% \\
nld & Latn & FineWeb2-HQ & 1.17\% \\
pol & Latn & FineWeb2-HQ & 1.15\% \\
ind & Latn & FineWeb2-HQ & 0.94\% \\
vie & Latn & FineWeb2-HQ & 0.80\% \\
kor & Hang & FineWeb2 & 0.76\% \\
tur & Latn & FineWeb2-HQ & 0.66\% \\
fas & Arab & FineWeb2-HQ & 0.62\% \\
ces & Latn & FineWeb2-HQ & 0.56\% \\
swe & Latn & FineWeb2-HQ & 0.56\% \\
ron & Latn & FineWeb2 & 0.55\% \\
arb & Arab & FineWeb2-HQ & 0.51\% \\
nob & Latn & FineWeb2 & 0.50\% \\
hun & Latn & FineWeb2-HQ & 0.48\% \\
dan & Latn & FineWeb2-HQ & 0.44\% \\
ukr & Cyrl & FineWeb2 & 0.40\% \\
tha & Thai & FineWeb2 & 0.39\% \\
ell & Grek & FineWeb2-HQ & 0.36\% \\
fin & Latn & FineWeb2 & 0.32\% \\
hin & Deva & FineWeb2 & 0.18\% \\
ben & Beng & FineWeb2 & 0.10\% \\
zsm & Latn & FineWeb2 & 0.09\% \\
\bottomrule
\end{tabular}
\end{table} 

\section{Details on Contrastive Training}
\label{appendix:contrastive-details}
The data used for each of our training stages differ in format, size, and language distribution. To improve the quality of in-batch negatives, a single batch is made up of data from a single dataset; the target dataset is selected randomly with probability proportional to the size of the dataset.
\paragraph{Pair Training.}
Our models are trained using contrastive learning with InfoNCE loss, where each query uses all in-batch negative documents and all other queries as negatives, with a temperature of 0.02.
Critically, we apply INT8 tanh quantization from the beginning of training, enabling the models to learn representations optimized for reduced precision.
Both \pplx-0.6B and \pplx-4B are trained with a global batch size of 16,384 for 50,000 steps at a sequence length of 256 tokens.
We use learning rates of~$2\times 10^{-4}$ and~$5 \times 10^{-5}$ for the 0.6B and 4B models, respectively, selected to balance training stability and convergence speed across different model scales.

\paragraph{Contextual Training.}
We train both models initialized from the checkpoint obtained after pair training with quantization applied during both training and inference. The training corpus comprises four contextual retrieval datasets: synthetic MLDR (LLM-generated queries), MLDR, NarrativeQA, and SQuAD, all formatted with chunk-level annotations for contextual embedding learning. Additionally, we synthesize queries for MLDR chunks lacking associated queries. During synthesis, we incorporate neighborhood information into our prompts to prevent generating queries relevant to multiple chunks. Documents are partitioned into chunks of 256 tokens with a fixed number of 16 chunks per document.

We employ a hybrid Matryoshka~\citep{kusupati2022matryoshka} dual-objective loss that combines global document-level InfoNCE with a local loss, training in six embedding dimensions [128, 256, 512, 1024, 2048, 2560] with equal weighting. The dual objective weight~$\beta$ is scheduled from 0.2 to 0.5 during training using a cosine annealing schedule. To mitigate false negatives, we implement three masking strategies: relative similarity thresholding to mask negatives within 0.1 of the positive similarity, duplicate document masking, and query-query false negative masking where additional query-query similarities serve as hard negatives for the global loss. Training runs for 7,000 steps with AdamW optimization (learning rate~$10^{-5}$, weight decay~$0.1$) and gradient clipping at 1.0. The 0.6B and 4B models are trained with batch sizes of 128 and 16, respectively. The temperature is set to 0.02.

\paragraph{Triplet Training.}
We perform triplet training using an InfoNCE loss. In this stage, each query uses in-batch negative documents augmented with 3 mined hard negatives.
We increase the sequence length to 512 tokens and use a global batch size of 512 with gradient accumulation of 4.
The temperature is set to 0.03.
Both models are fine-tuned for 2,000 steps, with learning rates of~$5 \times 10^{-5}$ and~$10^{-5}$ for the 0.6B and 4B models, respectively.

\section{Details on MTEB Evaluation}
\label{appendix:mteb-eval-details}
We provide per-task scores for the MTEB(Multilingual, v2) retrieval benchmark in Table~\ref{tab:mmteb-per-task} and per-task scores for the MTEB(Code) benchmark in Table~\ref{tab:mteb-code-per-task}. 
We evaluate \pplx with a sequence length of 1024 on all tasks except LEMBPasskeyRetrieval, where we use a sequence length of 16,384. 
We observe that the SyntheticText2SQL task contains duplicate documents in its corpus. To break the symmetry between these duplicates, we inject small random noise into our embeddings.
Since the ArguAna task requires the embedding model to find \emph{refuting} documents instead of supporting ones, we find that prepending queries for this task with the string \texttt{"Given a claim, find documents that refute the claim. Claim: "} substantially improves performance. Due to the special structure of ArguAna, it is the only dataset where we apply a prompt.

\begin{table}[htp]
    \centering
    \caption{nDCG@10 on MTEB(Multilingual, v2) retrieval tasks}
    \label{tab:mmteb-per-task}
    \resizebox{\textwidth}{!}{\begin{tabular}{l|ccccc|ccc}
\toprule
Task & \rotatebox{90}{qwen3-embed-4B} & \rotatebox{90}{text-embedding-3-large} & \rotatebox{90}{gemini-embedding-001} & \rotatebox{90}{\pplx-4B (\textbf{INT8})} & \rotatebox{90}{\pplx-4B (\textbf{BIN})} & \rotatebox{90}{qwen3-embed-0.6B} & \rotatebox{90}{\pplx-0.6B (\textbf{INT8})} & \rotatebox{90}{\pplx-0.6B (\textbf{BIN})} \\
\midrule
Average & \underline{69.60} & 59.27 & 67.71 & \textbf{69.66} & 68.22 & \underline{64.65} & \textbf{65.41} & 61.44 \\
\midrule
AILAStatutes & \textbf{81.19} & 41.85 & 48.77 & \underline{61.00} & 60.26 & \textbf{79.02} & \underline{59.84} & 53.77 \\
ArguAna & \underline{75.64} & 57.99 & \textbf{86.44} & 69.99 & 67.81 & \textbf{70.96} & \underline{65.94} & 61.07 \\
BelebeleRetrieval & \underline{81.16} & 68.79 & \textbf{90.73} & 77.88 & 76.74 & 68.74 & \textbf{72.86} & \underline{68.99} \\
CovidRetrieval & \textbf{87.37} & 68.43 & \underline{79.13} & 77.50 & 76.98 & \textbf{84.76} & \underline{77.37} & 74.35 \\
HagridRetrieval & 98.77 & \underline{99.04} & \textbf{99.31} & 98.77 & 98.77 & 98.76 & \textbf{98.77} & \textbf{98.77} \\
LEMBPasskeyRetrieval & \underline{84.25} & 69.75 & 38.50 & \textbf{89.50} & 79.25 & \textbf{84.75} & \underline{76.75} & 60.25 \\
LegalBenchCorporateLobbying & \underline{95.42} & 95.22 & \textbf{95.98} & 94.85 & 93.78 & \textbf{94.52} & \underline{94.36} & 93.01 \\
MIRACLRetrievalHardNegatives & \underline{69.49} & 56.94 & \textbf{70.42} & 66.24 & 64.64 & 61.23 & \textbf{68.56} & \underline{64.21} \\
MLQARetrieval & 81.92 & 73.25 & \textbf{84.16} & \underline{83.34} & 81.77 & 72.79 & \textbf{78.98} & \underline{74.83} \\
SCIDOCS & \textbf{31.44} & 23.07 & \underline{25.15} & 23.87 & 23.33 & \textbf{24.41} & \underline{22.83} & 21.70 \\
SpartQA & 20.15 & 7.44 & 10.30 & \textbf{23.30} & \underline{22.04} & \textbf{10.58} & \underline{9.17} & 7.60 \\
StackOverflowQA & \underline{94.32} & 92.44 & \textbf{96.71} & 93.61 & 93.08 & \underline{89.99} & \textbf{90.65} & 88.97 \\
StatcanDialogueDatasetRetrieval & 42.28 & 31.10 & 51.11 & \textbf{56.53} & \underline{55.56} & 33.63 & \textbf{41.14} & \underline{34.79} \\
TRECCOVID & \textbf{92.92} & 79.56 & \underline{86.32} & 85.61 & 83.94 & \textbf{90.52} & \underline{85.90} & 81.33 \\
TempReasonL1 & 1.23 & \underline{2.13} & \textbf{2.96} & 1.19 & 1.16 & 1.02 & \textbf{1.43} & \underline{1.40} \\
TwitterHjerneRetrieval & 72.58 & \underline{81.44} & \textbf{98.02} & 75.67 & 75.24 & 60.04 & \textbf{70.04} & \underline{65.20} \\
WikipediaRetrievalMultilingual & 91.23 & 89.24 & \textbf{94.20} & \underline{92.68} & 92.30 & 87.13 & \textbf{90.98} & \underline{89.74} \\
WinoGrande & 51.51 & 29.11 & 60.52 & \textbf{82.40} & \underline{81.28} & 50.79 & \textbf{71.85} & \underline{65.92} \\
\bottomrule
\end{tabular}}
\end{table}

\begin{table}[htp]
    \centering
    \caption{nDCG@10 on MTEB(Code) retrieval tasks}
    \label{tab:mteb-code-per-task}
    \resizebox{\textwidth}{!}{\begin{tabular}{l|ccccc|ccc}
\toprule
Task & \rotatebox{90}{qwen3-embed-4B} & \rotatebox{90}{text-embedding-3-large} & \rotatebox{90}{gemini-embedding-001} & \rotatebox{90}{\pplx-4B (\textbf{INT8})} & \rotatebox{90}{\pplx-4B (\textbf{BIN})} & \rotatebox{90}{qwen3-embed-0.6B} & \rotatebox{90}{\pplx-0.6B (\textbf{INT8})} & \rotatebox{90}{\pplx-0.6B (\textbf{BIN})} \\
\midrule
Average & \textbf{80.07} & 66.54 & 76.00 & \underline{78.73} & 78.11 & \underline{75.42} & \textbf{75.85} & 73.91 \\
\midrule
AppsRetrieval & \underline{89.18} & 28.46 & \textbf{93.75} & 87.81 & 86.65 & \underline{75.34} & \textbf{78.66} & 71.70 \\
COIRCodeSearchNetRetrieval & \textbf{87.93} & 75.54 & 81.06 & \underline{84.39} & 83.70 & \textbf{84.69} & \underline{82.38} & 80.15 \\
CodeEditSearchRetrieval & 76.49 & 71.11 & \textbf{81.61} & \underline{81.26} & 80.23 & 64.42 & \textbf{76.36} & \underline{72.88} \\
CodeFeedbackMT & \textbf{93.21} & 68.92 & 56.28 & \underline{86.17} & 85.67 & \textbf{90.82} & \underline{82.96} & 81.79 \\
CodeFeedbackST & \textbf{89.51} & 80.42 & 85.33 & \underline{86.95} & 86.52 & \textbf{86.39} & \underline{84.73} & 83.26 \\
CodeSearchNetCCRetrieval & \textbf{95.59} & 73.18 & 84.69 & \underline{92.27} & 91.67 & \textbf{91.72} & \underline{88.10} & 86.04 \\
CodeSearchNetRetrieval & \textbf{92.34} & 90.50 & \underline{91.33} & 90.81 & 90.54 & \textbf{91.01} & \underline{89.86} & 88.73 \\
CodeTransOceanContest & \textbf{90.99} & 84.25 & \underline{89.53} & 88.38 & 88.61 & \textbf{86.05} & \underline{85.03} & 84.86 \\
CodeTransOceanDL & \textbf{35.04} & \underline{34.23} & 31.47 & 33.91 & 33.11 & 31.36 & \underline{35.16} & \textbf{35.36} \\
CosQA & 37.98 & 31.00 & \textbf{50.24} & \underline{42.34} & 41.01 & 36.48 & \textbf{43.32} & \underline{41.04} \\
StackOverflowQA & \underline{94.32} & 92.44 & \textbf{96.71} & 93.61 & 93.08 & \underline{89.99} & \textbf{90.65} & 88.97 \\
SyntheticText2SQL & \textbf{78.21} & 68.45 & 69.96 & \underline{76.80} & 76.52 & \textbf{76.74} & \underline{72.98} & 72.18 \\
\bottomrule
\end{tabular}}
\end{table}

\section{Details on ConTEB Evaluation}
\label{appendix:conteb-eval-details}
Our ConTEB evaluation process employs two distinct contextual embedding strategies depending on document characteristics. For standard-length documents, we use the \texttt{ContextualEmbedder}, which generates contextual embeddings by incorporating surrounding chunk information during encoding. For exceptionally long documents (notably the ESG Reports dataset, which contains documents exceeding 30,000 tokens), we implement the \texttt{FixedContextualEmbedder} with a fixed partitioning strategy that divides documents into overlapping segments of configurable size (default: 10,000 tokens with 35-chunk overlap).

\section{Details on BERGEN Evaluation}
\label{appendix:bergen-eval-details}

\paragraph{Embedding Models.} Within the BERGEN framework, we implement a custom retriever class that uses the official sentence-transformers implementations of \pplx, Qwen3-Embedding, and BGE-M3, ensuring fair evaluations. When encoding queries, we set \texttt{prompt\_name="query"} for the Qwen3-Embedding models.

\paragraph{RAG Configuration.} Below, we provide the command we use for performing the end-to-end RAG evaluations. While the top-100 passages are retrieved from the KILT dump, only the top-5 passages are presented to the generator.
\begin{verbatim}
python3 -u bergen.py retriever="${retriever}" \ 
    generator="vllm_qwen-25-32b-instruct" \
    dataset="${dataset}" retrieve_top_k=100 \
    generator.init_args.max_length=32768
\end{verbatim}

\end{document}